\documentclass[sigconf]{acmart}

\AtBeginDocument{%
  }

\usepackage{multirow}
\usepackage{xcolor}            
\usepackage{soul}              
\usepackage{placeins}          
\usepackage{multirow}          
\usepackage{xspace}
\usepackage{subcaption}
\usepackage{booktabs}
\usepackage{array}

\newcommand{\hlc}[2][yellow]{{%
  \colorlet{foo}{#1}%
  \sethlcolor{foo}\hl{#2}}%
}

\definecolor{ExperiencedColor}{RGB}{255,204,153}
\definecolor{InternalizedColor}{RGB}{204,229,255}
\definecolor{AnticipatedColor}{RGB}{255,153,153}
\definecolor{StructuralColor}{RGB}{255,204,229}
\definecolor{InformationSColor}{RGB}{204,255,229}    
\definecolor{EmotionalSColor}{RGB}{255,229,204}      
\definecolor{EsteemSColor}{RGB}{229,204,255}         
\definecolor{TangibleSColor}{RGB}{255,255,204}       
\definecolor{GroupSColor}{RGB}{204,255,255}          
\definecolor{WhiteColor}{RGB}{255,255,255}      

\newcommand{\corpus}{\textsc{SCOPE}\xspace}

\usepackage{float}

\copyrightyear{2026}
\acmYear{2026}
\setcopyright{cc}
\setcctype{by}
\acmConference[HT '26]{37th ACM Conference on Hypertext}{September 14--18, 2026}{London, United Kingdom}
\acmBooktitle{37th ACM Conference on Hypertext (HT '26), September 14--18, 2026, London, United Kingdom}
\acmDOI{10.1145/3800935.3830853}
\acmISBN{979-8-4007-2564-7/2026/09}
\begin{document}

\title{Stigma and Support in Online Sexual Violence Narratives on Reddit}

\author{Shirlene Rose Bandela}
\affiliation{%
  \institution{Virginia Tech}
  \city{Alexandria}
  \country{USA}}
\email{shirleneroseb@vt.edu}
\author{Karan Bindal}
\affiliation{%
  \institution{Drexel University}
  \city{Philadelphia}
  \country{USA}}
\email{kb3887@drexel.edu}
\author{Vaibhav Garg}
\authornote{Both authors contributed equally.}
\affiliation{%
  \institution{Virginia Tech}
  \city{Alexandria}
  \country{USA}}
\email{vaibhavg@vt.edu}
\author{Rezvaneh Rezapour}
\authornotemark[1]
\affiliation{%
  \institution{Drexel University}
  \city{Philadelphia}
  \country{USA}}
\email{sr3563@drexel.edu}

\begin{abstract}
\textcolor{blue}{\textit{Warning: This paper discusses sexual violence and may contain material that some readers, particularly survivors, may find distressing.}}

\noindent Online communities increasingly provide spaces where survivors of sexual violence can share their experiences and seek support. Although prior research has examined stigma and social support separately, less is known about how stigma expressed in survivor narratives relates to the support offered in response. We introduce the \corpus dataset, linking stigma signals in online survivor narratives to support types in corresponding comment threads. We annotate posts using a multi-dimensional stigma taxonomy, including \textit{Experienced, Internalized, Anticipated,} and \textit{Structural Stigma}, and comments using a support taxonomy encompassing \textit{Information Support, Emotional Support, Esteem Support, Tangible Assistance,} and \textit{Group Interaction}. Using contextual, linguistic, and emotion analyses, we compare \textit{Stigma} and \textit{No Stigma} content and find that \textit{Stigma} narratives place greater emphasis on internalized distress, whereas \textit{No Stigma} narratives focus more on interpreting situations and experiences. \textit{Internalized Stigma} is the most prevalent category, and community responses remain broadly stable across stigma types, with \textit{Information} and \textit{Esteem Support} appearing most often. These findings show how stigma shapes survivor narratives and peer responses and have implications for computational modeling, content moderation, and safer online systems.
\end{abstract}




\maketitle

\section{Introduction} \label{sec:ref}
Online platforms such as Reddit have become important socio-technical spaces where individuals share lived experiences and seek advice or support from peers \cite{o2018today, andalibi2016understanding, alaggia2020never, bair2022reddit, bouzoubaa-etal-2024-decoding}. While these interactions can foster support and solidarity, survivor narratives may also reveal the complex ways in which stigma is experienced, anticipated, and internalized, as well as how communities respond to these experiences \cite{goffmanStigmaNotesManagement1968, kennedy2018still, stangl2019health, lanthier2023coming, link2001conceptualizing}. Understanding how stigma manifests in survivor narratives and how online communities respond is therefore critical to designing safer digital spaces and more supportive interventions \cite{lanthier2023coming, andalibi2018social}. 

Prior research has largely examined stigma and social support separately. Studies of stigma have analyzed stigmatizing language in online narratives \citep{bouzoubaa2025phenotypes, giorgi2024lived} and investigated how narrative framing relates to stigma in mental health discussions \citep{mittal2023moral,bouzoubaa2026cognitive}. In parallel, research on online social support has identified distinct forms of support and examined how they are expressed and exchanged within online communities \citep{de2014mental, de2017language, sharma2018mental, sharma2020engagement}. However, limited work has jointly examined how stigma expressed in a post relates to the support provided in response, particularly across fine-grained stigma categories within online communities for survivors of sexual violence.

To address this gap, we introduce \corpus\ (\textbf{S}tigma and \textbf{CO}mmunity \textbf{P}eer \textbf{E}xpressions), a novel dataset of Reddit posts about sexual violence and their associated comment threads, annotated for stigma and community support. Posts are labeled using a multilevel stigma taxonomy that distinguishes \textit{Stigma} from \textit{No Stigma} and further categorizes stigma as \textit{Experienced, Internalized, Anticipated, or Structural} \cite{bouzoubaa2024words}. Comments are annotated using the support taxonomy developed by \citet{lopes2019classification}, which captures \textit{Information, Emotional, Esteem, Tangible}, and \textit{Group-Interaction} support. By linking survivor disclosures with community responses, our framework enables systematic study of how different forms of stigma relate to peer-support strategies.

Our analyses show that posts labeled \textit{Stigma} emphasize validation, whereas posts labeled as \textit{No Stigma} lean toward informational and guidance-oriented support; however, this distinction is not absolute, as support strategies remain largely consistent across both labels, differing mainly in emphasis. At the post level, LIWC analysis reveals that \textit{Internalized, Experienced}, and \textit{Anticipated Stigma} exhibit higher \textit{Authenticity} but lower \textit{Analytic} scores than \textit{Structural Stigma}, reflecting more personal, subjective narratives, while posts labeled as \textit{No Stigma} fall between these groups. At the comment level, \textit{Esteem Support} exhibits the highest \textit{Clout}, \textit{Group Interaction} exhibits the highest \textit{Authenticity}, and \textit{Information Support} shows the highest \textit{Analytic} scores, aligning with their respective communicative roles. Emotion analysis also indicates greater emotional intensity in self-blaming narratives than in \textit{Structural Stigma} and \textit{No Stigma} content.
These findings demonstrate how stigma shapes survivor narratives and peer responses and highlight implications for computational modeling, content moderation, and the design of safer online environments. We release the annotated dataset and code at \url{https://github.com/ShirleneRose/Stigma_SV}.

\begin{table}[t]
\centering
\resizebox{0.9\linewidth}{!}{
\begin{tabular}{p{0.5cm} p{2cm} p{7cm}}
\toprule
&\textbf{Category} & \textbf{Definition} \\
\midrule

\multirow{4}{*}{\rotatebox{90}{\textbf{Stigma}}}
&Experienced & Direct stigma from others, e.g., blame, disbelief, or dismissal
\\

&Internalized & Self-directed stigma, such as shame, guilt, or self-blame 
\\

&Anticipated &Fear or expectation of judgment, blame, or negative reactions from others 
\\

&Structural &Stigma embedded in institutional or societal systems, such as legal or cultural barriers
\\

\midrule

\multirow{5}{*}{\rotatebox{90}{\textbf{Support}}}
&Information Support & Advice, information, or guidance about resources or next steps
\\

&Emotional Support &Expressions of empathy or compassion
\\

&Esteem Support & Affirmation of worth, validation, or encouragement
\\

&Tangible Assistance & Offers of concrete help or direct support
\\

&Group Interaction &References to shared experiences or community belonging 
\\

\bottomrule
\end{tabular}}
\caption{Definitions of stigma and support categories used in the annotation task.}
\label{tab:category_definitions}
\vspace{-0.5cm}
\end{table}

\section{Modeling Stigma and Support in Survivor Narratives}
\label{sec:stigma_support_narratives}
\noindent\textbf{Stigma in Survivor Narratives. } 
Stigma is commonly understood as a socially discrediting attribute that shapes how individuals are perceived and treated \cite{goffmanStigmaNotesManagement1968}. Stigma is also framed as a multi-level social process embedded in cultural norms, institutional practices, and power dynamics \citep{link2001conceptualizing, aggleton2003stigma}. 
Prior work identifies several key mechanisms of stigma, including enacted (experienced discrimination), anticipated (expectations of judgment), and internalized (self-directed stigma) \citep{earnshaw2009conceptualizing, corrigan2005stigma}, as well as structural stigma, which reflects institutional and systemic barriers \citep{stangl2019health, hatzenbuehler2013stigma}. 
In the context of sexual violence, these forms of stigma often co-occur in narratives; survivors may recount blame from others, express fear of disclosure, or articulate feelings of shame and distrust toward institutions \cite{kennedy2018still, ullman2023talking}.
Building on these frameworks, we adopt a taxonomy comprising \textbf{\textit{Experienced, Internalized, Anticipated, and Structural Stigma}} \cite{bouzoubaa2025phenotypes}, as these categories capture distinct dimensions of stigma observable in textual narratives. 

\noindent\textbf{Support in Community Responses. }
To characterize how communities respond to survivor narratives, we draw on an established framework of online social support \citep{lopes2019classification}. This framework identifies functional types of support commonly observed in online health and crisis contexts \citep{de2014mental, sharma2020engagement}. 
We consider \textbf{\textit{Information Support}} (advice or guidance), \textbf{\textit{Emotional Support}} (expressions of empathy and care), \textbf{\textit{Esteem Support}} (affirmation of worth or validation), \textbf{\textit{Tangible Assistance}} (offers of direct or practical help), and \textbf{\textit{Group Interaction}} (signals of shared experience or belonging). 
These categories capture distinct functions that are particularly relevant to discussions of sexual violence, where survivors may seek validation, reassurance, or actionable guidance. For example, \textit{Emotional Support} can validate experiences, while \textit{Information Support} may guide next steps, underscoring the importance of distinguishing between these categories in our analysis.  
Table~\ref{tab:category_definitions} summarizes the stigma and support categories and their definitions.

To illustrate these categories, we present two paraphrased examples from Reddit, highlighting color-coded excerpts that reflect specific stigma and support types. Example 1 shows multiple co-occurring forms of stigma within a single narrative, while Example 2 demonstrates several forms of community support responses.

\begin{figure}[h]
\begingroup
\setlength{\fboxrule}{0.5pt}
\setlength{\fboxsep}{6pt}

\noindent
\fcolorbox{black!60}{white}{%
  \begin{minipage}{%
    \dimexpr\linewidth-2\fboxsep-2\fboxrule\relax
  }
  \scriptsize

  \textbf{Stigma Categories:}
  \hlc[ExperiencedColor]{Experienced Stigma}
  \hlc[InternalizedColor]{Internalized Stigma}
  \hlc[AnticipatedColor]{Anticipated Stigma}
  \hlc[StructuralColor]{Structural Stigma}

  \medskip
  \textbf{Example 1: Stigma Excerpts from a Sexual Violence Narrative}

  \smallskip
  \hlc[StructuralColor]{I've been trying to get someone to see my story for six plus months with no success}
  \hlc[WhiteColor]{...}
  \hlc[ExperiencedColor]{He threatened to murder me and make an example of me for coming forward}
  \hlc[WhiteColor]{...}
  \hlc[AnticipatedColor]{I feared other forms of negative attention}
  \hlc[WhiteColor]{...}
  \hlc[InternalizedColor]{I just feel so embarrassed and ashamed like I did something wrong}
  \hlc[WhiteColor]{...}

  \medskip
  \textbf{Support Categories:}
  \hlc[InformationSColor]{Information Support}
  \hlc[EmotionalSColor]{Emotional Support}
  \hlc[EsteemSColor]{Esteem Support}
  \hlc[TangibleSColor]{Tangible Assistance}
  \hlc[GroupSColor]{Group Interaction}

  \medskip
  \textbf{Example 2: Support Excerpts from Responses to a Sexual Violence Narrative}

  \smallskip
  \hlc[EmotionalSColor]{I am so sorry to hear about this!}
  \hlc[WhiteColor]{...}
  \hlc[InformationSColor]{I'd recommend reaching out to RAINN. It's a free hotline.}
  \hlc[WhiteColor]{...}
  \hlc[EsteemSColor]{You are incredibly brave for sharing your story}
  \hlc[WhiteColor]{...}
  \hlc[TangibleSColor]{DM me if you'd like to talk more about that}
  \hlc[WhiteColor]{...}
  \hlc[GroupSColor]{We hear you. We believe you. And you now have /all/ of us in your corner.}
  \hlc[WhiteColor]{...}
  \end{minipage}%
}
\endgroup
\end{figure}

\section{Methodology}
\subsection{Data Collection}
\label{sec:reddit_stories}

We built on a dataset of sexual violence narratives curated by \citet{garg2025analyzing}, which includes 5,328 posts from three subreddits: \textit{r/meToo}, \textit{r/SexualHarassment}, and \textit{r/sexualassault}. The dataset's domain specificity and relevance to survivor disclosures make it well-suited to our analysis.
To capture community responses, we collected direct replies to these posts using Python's Reddit API Wrapper (PRAW).
We performed standard preprocessing by removing empty posts and comments as well as those labeled as ``[deleted]'' or ``[removed].'' We also excluded posts without comments to ensure that each post had an interactional context for analyzing support. The final dataset includes 3,675 posts and 5,131 comments.

\subsection{Data Annotation}
\subsubsection{Annotating Posts}
We adopted a multi-stage annotation pipeline to identify and characterize stigma in Reddit discussions of sexual violence. Posts were first screened for relevance, then labeled as \textit{Stigma} or \textit{No Stigma}, and finally annotated at a fine-grained level to capture specific types of stigma. Below, we describe each stage.

\noindent\textbf{Relevance Checking.} We screened posts for topical relevance and labeled them as `Relevant' if they directly discussed sexual violence, and `Not Relevant' otherwise.

\noindent\textbf{Stigma vs. No Stigma.} After filtering for relevance, posts were assigned a binary label: \textit{Stigma} if they contained stigmatizing language, beliefs, or experiences, and \textit{No Stigma} otherwise.

\noindent\textbf{Fine-grained Stigma Labeling.} We applied fine-grained annotation using the taxonomy in Table~\ref{tab:category_definitions}, labeling posts as \textit{Experienced}, \textit{Internalized}, \textit{Anticipated}, and/or \textit{Structural Stigma}, allowing multiple labels per post. Annotation was conducted iteratively, with disagreements resolved through discussion and guideline refinement until consensus was reached. As shown in Table~\ref{tab:stigma_kappa}, agreement improved across rounds, with initially subjective categories (e.g., \textit{Internalized Stigma}, \textit{Anticipated Stigma}) reaching substantial agreement over time.

\subsubsection{Annotating Support in Comments} We annotated comments for support types using the taxonomy in Table~\ref{tab:category_definitions}, allowing multiple labels per comment. Annotation was conducted iteratively with disagreements resolved through discussion and guideline refinement. As shown in Table~\ref{tab:support_kappa}, agreement was generally high, with more explicit categories (e.g., \textit{Information Support}) showing stronger consistency and others (e.g., \textit{Esteem Support}) improving over time.

\begin{table}[t]
\centering
\normalsize
\resizebox{0.7\linewidth}{!}{
\begin{tabular}{lccc}
\toprule
\textbf{Tag} & Round 1 & Round 2 & Round 3 \\
\midrule
Anticipated    & 0.407 & 0.510 & 0.608 \\
Experienced    & 0.268 & 0.675 & 0.722 \\
Internalized   & 0.396 & 0.558 & 0.639 \\
Structural     & 0.375 & 0.733 & 0.852 \\
No Stigma      & 0.385 & 0.610 & 0.834 \\
Not Applicable & 0.508 & 0.643 & 1.000 \\
\bottomrule
\end{tabular}
}
\caption{Cohen's kappa scores for \textit{Stigma} categories ($\kappa$).}
\label{tab:stigma_kappa}
\end{table}
\begin{table}[t]
\centering
\normalsize
\resizebox{0.7\linewidth}{!}{
\begin{tabular}{lccc}
\toprule
\textbf{Tag} & Round 1 & Round 2 & Round 3 \\
\midrule
Information Support & 0.690 & 0.639 & 0.970 \\
Emotional Support   & 0.884 & 0.871 & 0.821 \\
Esteem Support      & 0.542 & 0.795 & 0.778 \\
Group Interaction   & 0.831 & 0.901 & 0.838 \\
Tangible Assistance & 0.718 & 0.901 & 1.000 \\
None                & 0.918 & 0.846 & 1.000 \\
\bottomrule
\end{tabular}
}
\caption{Cohen's kappa scores for support categories ($\kappa$).}
\label{tab:support_kappa}
\end{table}

\subsection{LLM-Based Classification and Scaling of Stigma and Support Annotations}
To extend the annotated data, we applied a $K$-shot in-context learning approach to classify stigma in posts and support in comments. Each category was treated as a separate binary yes/no decision, allowing multi-label assignments.

\noindent\textbf{Stigma Classification.}
We first screened posts for relevance, labeling them as `Relevant' if they directly discussed sexual violence. Relevant posts were then assigned a coarse label of \textit{Stigma} or \textit{No Stigma}. 
For fine-grained stigma classification, each post was paired with the definition of a single target category (\textit{Experienced}, \textit{Internalized}, \textit{Anticipated}, or \textit{Structural}) along with $K=5$ semantically similar labeled examples. We chose $K=5$ based on pilot experiments to balance contextual diversity with prompt length. Examples were retrieved using cosine similarity over MPNet-based SentenceTransformer embeddings \cite{MPNET-SentenceTransformer}, ensuring semantic relevance and stylistic similarity.
The model followed a fixed procedure: it read the post, compared it to the category definition, and decided whether the stigma type was present (yes/no). We set the temperature to 0 to ensure consistency \citep{Prompt-Engineering2025}. After validating this setup on the annotated subset, we applied it to the remaining data, achieving strong alignment with human labels and enabling scalable, consistent annotation.

\noindent\textbf{Support Classification. }
We applied the same in-context learning approach to support classification. Each comment was evaluated against one category at a time (\textit{Information}, \textit{Emotional}, \textit{Esteem}, \textit{Tangible}, \textit{Group Interaction}) using $K=5$ semantically similar examples retrieved via MPNet-based cosine similarity. The prompt mirrored the stigma classification setup, combining the category definition, examples, and a binary yes/no decision.
We set the temperature to 0 for consistency. After validating the approach on the annotated subset, we applied it to the remaining comments, enabling efficient annotation while maintaining alignment with the taxonomy.

\subsection{Contextual Analysis }

\noindent\textbf{Topic Analysis.}
\label{sec:LLOOM}
To identify recurring conceptual patterns in both sexual violence narratives and community responses, we applied the LLooM framework \cite{lam2024concept}, a large language model–assisted method for extracting higher-level concepts from text collections. LLooM combines concept generation and example matching using LLMs to surface interpretable themes that capture common patterns across posts and comments.

\noindent\textbf{LIWC.} To analyze linguistic patterns in narratives and responses, we used Linguistic Inquiry and Word Count (LIWC) \cite{boyd2022development}, which captures linguistically meaningful features such as emotional expression, cognitive processes, and social language.

\noindent\textbf{Emotion Analysis.} To quantify emotional expression, we used the NRC Emotion Lexicon \cite{mohammad2013crowdsourcing} to capture eight core emotions. We compared emotion distributions across \textit{Stigma} and \textit{No Stigma} posts and their corresponding comments, highlighting differences in emotional tone and community responses.

\section{Results}
\subsection{Classification Results}

\noindent\textbf{Stigma Categories. }
\label{sec:model_evaluation}
Table~\ref{tab:full_macro_results} presents the macro-averaged precision, recall, and F1 scores for each of the models. 
Level 1 classifies posts as `Relevant' or `Not Relevant,' Level 2 distinguishes \textit{Stigma} from \textit{No Stigma}, and Level 3 assigns fine-grained stigma categories.
At Level 1, Gemini (2.0 Flash) achieves the best performance (F1 = 0.957), with a more balanced precision–recall trade-off than other models. At Level 2, Gemini again achieves the highest F1 score (F1 = 0.820), while GPT-4o-mini shows higher precision but lower recall, resulting in a lower overall F1. Llama and GPT-oss achieve lower F1 scores when distinguishing \textit{Stigma} from \textit{No Stigma}.
At Level 3, which is a multi-label classification task, Gemini maintains the strongest performance (F1 = 0.773). Llama and GPT favor recall over precision, while GPT-oss is more conservative, favoring precision over recall.

\begin{table}[t]
\centering
\small
\setlength{\tabcolsep}{4pt}
\resizebox{\linewidth}{!}{%
\begin{tabular}{lccccccccc}
\toprule
& \multicolumn{3}{c}{\textbf{Rel. vs Not Rel.}} 
& \multicolumn{3}{c}{\textbf{Stig. vs No Stig.}} 
& \multicolumn{3}{c}{\textbf{Fine-grained Stig.}} \\
\cmidrule(lr){2-4} \cmidrule(lr){5-7} \cmidrule(lr){8-10}
\textbf{Model} & \textbf{P} & \textbf{R} & \textbf{F1} & \textbf{P} & \textbf{R} & \textbf{F1} & \textbf{P} & \textbf{R} & \textbf{F1} \\
\midrule
gemini-2.0-flash & 0.957 & 0.957 & \textbf{0.957} & 0.820 & 0.820 & \textbf{0.820} & 0.732 & 0.825 & \textbf{0.773} \\
Llama-3.3-70B    & 0.808 & 0.638 & 0.684 & 0.711 & 0.823 & 0.727 & 0.593 & 0.883 & 0.683 \\
gpt-4o-mini      & 0.854 & 0.709 & 0.760 & 0.891 & 0.729 & 0.777 & 0.570 & 0.924 & 0.683 \\
gpt-oss-120b     & 0.670 & 0.869 & 0.716 & 0.700 & 0.829 & 0.680 & 0.776 & 0.562 & 0.652 \\
\bottomrule
\end{tabular}}
\caption{Macro Precision (P), Recall (R), and F1 across \textit{Stigma} classification tasks. Relevant (Rel.), Stigma (Stig.)}
\label{tab:full_macro_results}
\vspace{-0.5cm}
\end{table}

\begin{table}[t]
\centering
\resizebox{\linewidth}{!}{%
\begin{tabular}{lcccccc}
\toprule
\textbf{Stig. Type} & \textbf{\#Posts} & \textbf{} & \textbf{} & \textbf{\#Comments} & \textbf{} & \textbf{} \\
\cmidrule{3-7}
 & & \textbf{Emo.} & \textbf{Esteem} & \textbf{Group} & \textbf{Info.} & \textbf{Tangible} \\
\midrule
Internalized & 1809 & 988 & 1108 & 444 & 1326 & 228 \\
Anticipated & 1325 & 681 & 730 & 317 & 960 & 177 \\
Experienced & 1273 & 669 & 681 & 286 & 915 & 156 \\
Structural  & 325 & 154 & 142 & 71 & 213 & 46 \\
\bottomrule
\end{tabular}}
\caption{Distribution of support types across Stigma categories in posts. Stigma (Stig.), Emotional (Emo.), Informational (Info.)}
\label{tab:support_stigma_posts}
\vspace{-0.5cm}
\end{table}

\noindent\textbf{Support Categories. }
For support classification in comments, Gemini achieves the strongest overall performance, with an F1 score of 0.798. In contrast, GPT-4o-mini attains slightly higher precision (0.806), reflecting a more conservative prediction strategy, but substantially lower recall (0.649), indicating that it misses a considerable number of support instances. 

\noindent\textbf{Performance Considerations and Task Complexity.}
The observed performance (F1 $\approx$ 0.7--0.8) reflects the inherent difficulty of stigma and support classification. These tasks involve nuanced, context-dependent, and socially grounded categories that lack clear lexical boundaries. Stigma detection is particularly challenging due to annotator subjectivity and overlapping concepts, which can co-occur within the same narrative and introduce labeling ambiguity \cite{giorgi2024lived, bouzoubaa2025phenotypes}. Similarly, support classification is complicated by the frequent co-occurrence of multiple, semantically similar support categories within a single comment.
Prior work reports comparable performance ranges for both stigma classification \cite{gottipati2021exploring, bouzoubaa2025phenotypes, bouzoubaa2024words} and support detection \cite{ahani2026social}, suggesting that our results reflect the complexity of the tasks rather than solely the limitations of the models. 

\noindent\textbf{Overall Data Labeling.} Based on the classification results, we selected Gemini 2.0 Flash as the primary model for labeling both posts and comments in the full dataset.
Table~\ref{tab:support_stigma_posts} presents the frequency of each support category across stigma categories in posts. \textit{Internalized Stigma} is the most prevalent category and is associated with the highest volume of support responses. Across all stigma categories, \textit{Informational} and \textit{Esteem Support} are the most common forms of response, while \textit{Tangible Support} remains relatively rare.

\subsection{Contextual Analysis}
\label{sec:exploratory_analysis}

\begin{table*}[t]
\centering
\small
\begin{minipage}{0.48\textwidth}
\centering
\resizebox{\linewidth}{!}{%
\begin{tabular}{
>{\centering\arraybackslash}p{1cm} 
>{\centering\arraybackslash}p{3cm} 
p{7cm}
}
\toprule
\textbf{Category} & \textbf{Concept} & \textbf{Example} \\
\midrule

\multirow{5}{*}{\textbf{\rotatebox{90}{Stigma}}}
& Self-Blame \& minimization after SV
& \textit{I can't help but feel extremely guilty about this and that it's my fault. I don't know if I gave him the wrong message...and not telling him to stop...but I kept pushing him away} \\ \cmidrule(l){2-3}

& Self-blame \& self-silencing after SV
& \textit{am I blowing it out of proportion? was it actually just a mistake and should I forgive him?} \\ \cmidrule(l){2-3}

& Stigma \& fear of disclosure after SV
& \textit{I feel disgusted and ick when I'm with her. I felt guilty at that time for hating her so much... I played it off like it was nothing when it actually bothered me and wasn't able to talk to her about it.} \\ \cmidrule(l){2-3}

& Intimacy avoidance due to trauma
& \textit{I've stopped talking to him, mostly stayed in my room, and be really nervous when parents r not home, I've been scarred.} \\ \cmidrule(l){2-3}

& Secondary victimization by partner
& \textit{he was defensive… so I must have given it to him… he called me a psycho… he proceeded to explain to me that I came back and that I am crazy.} \\

\midrule

\multirow{5}{*}{\textbf{\rotatebox{90}{No Stigma}}}
& Recounting assault
& \textit{earlier today at school I experienced unwanted touching…} \\ \cmidrule(l){2-3}

& Trauma impact \& well-being
& \textit{this is a very complicated situation… my sister… told my mum that my dad sexually abused her for many years.} \\ \cmidrule(l){2-3}

& Seeking clarification on SV 
& \textit{I’m confused about whether I’m overreacting or if this is actually inappropriate.} \\ \cmidrule(l){2-3}

& Institutional navigation \& sexual harassment
& \textit{there is no hr, no one higher than him…} \\ \cmidrule(l){2-3}

& Ambiguous consent \& intoxication
& \textit{I don’t remember agreeing, and I’m confused about whether I actually consented while I was intoxicated.} \\

\bottomrule
\end{tabular}}
\caption{LLooM-derived post concepts and examples across \textit{Stigma} and \textit{No Stigm}a narratives.}
\label{tab:lloom_table_posts}
\end{minipage}%
\hfill
\begin{minipage}{0.48\textwidth}
\centering
\resizebox{\linewidth}{!}{%
\begin{tabular}{
>{\centering\arraybackslash}p{1cm} 
>{\centering\arraybackslash}p{3cm} 
p{7cm}
}
\toprule
\textbf{Category} & \textbf{Concept} & \textbf{Example} \\
\midrule

\multirow{5}{*}{\textbf{\rotatebox{90}{Stigma}}}
& Normalizing \& Validating trauma responses
& \textit{he's in the wrong, not you. you did tell him you weren't interested with something clearer than words -- your actions you described} \\ \cmidrule(l){2-3}

& Experience sharing \& advice
& \textit{as a person who had been the recipient of sexual harassment, I strongly suggest that you document everything} \\ \cmidrule(l){2-3}

& Empowering survivor identity
& \textit{you were 10. do not look at the thoughts or actions or behaviors of a ten year old ... you did nothing wrong} \\ \cmidrule(l){2-3}

& Resource signposting for survivors
& \textit{if your friend needs free legal advice and support, you can contact rightsofwomen.org.uk} \\ \cmidrule(l){2-3}

& Rejecting intoxication as excuse
& \textit{I wouldn't forgive him, but I'm not you. he says he was out of his head drunk, but it doesn't ring true} \\

\midrule

\multirow{5}{*}{\textbf{\rotatebox{90}{No Stigma}}}
& Affirming emotional support 
& \textit{it’s okay to dm me} \\ \cmidrule(l){2-3}

& Empathetic validation and affirmation 
& \textit{stay strong. stand high. if she calls the police then that’s on her} \\ \cmidrule(l){2-3}

& Defining assault
& \textit{more like sexual assault!! you have every right to your own body and your rules!} \\ \cmidrule(l){2-3}

& Encouraging professional help 
& \textit{I would suggest calling your local women's shelter, they have the resources} \\ \cmidrule(l){2-3}

& Supporting trauma processing
& \textit{sexual harassment definition: ``behavior characterized by the making of unwelcome and inappropriate sexual remarks or physical advances''} \\

\bottomrule
\end{tabular}}
\caption{LLooM-derived support concepts and examples across \textit{Stigma} and \textit{No Stigma} narratives.}
\label{tab:lloom_table_support}
\vspace{-0.25cm}
\end{minipage}
\end{table*}

\begin{figure}[t]
\centering

\begin{minipage}{0.7\linewidth}
\centering
\includegraphics[width=0.8\linewidth]{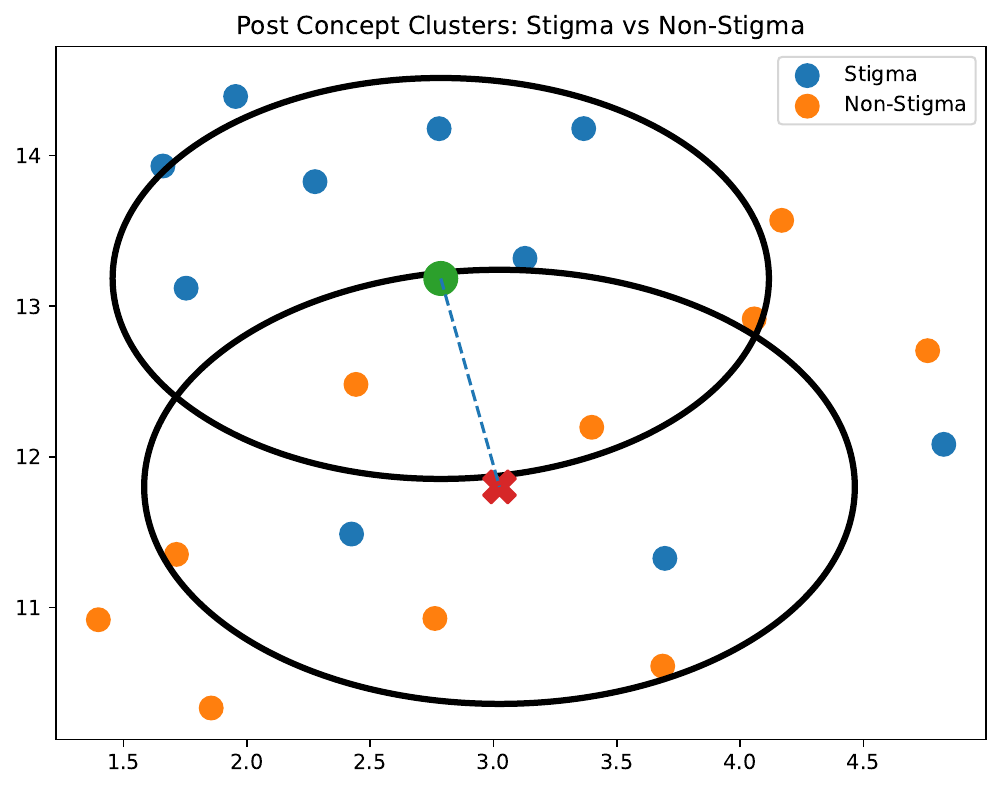}
\caption*{(a) \textit{Stigma} vs. \textit{No Stigma} post clusters.}
\end{minipage}
\vfill
\begin{minipage}{0.8\linewidth}
\centering
\includegraphics[width=0.7\linewidth]{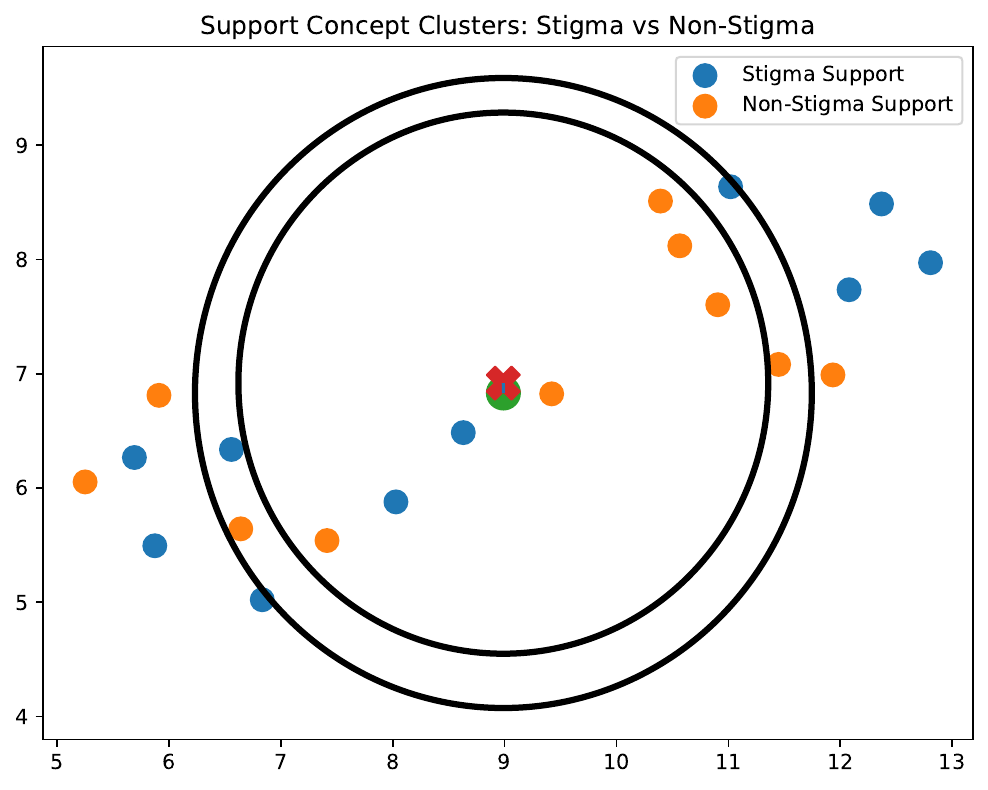}
\caption*{(b) Stigma vs. No Stigma support clusters.}
\end{minipage}

\caption{Cluster-based comparison of concepts across posts and comments.}
\label{fig:concept_clusters}
\vspace{-0.25cm}
\end{figure}

\noindent\textbf{Concept Discovery with LLooM.} Table~\ref{tab:lloom_table_posts} summarizes the most frequent concepts extracted from posts labeled as \textit{Stigma} and \textit{No Stigma}. \textit{Stigma}-related narratives are primarily characterized by themes of internalized distress and interpersonal harm, including internalized self-blame and minimization following sexual violence, internalized blame and self-silencing, fear of disclosure, and intimacy avoidance due to trauma. 
These themes reflect narratives centered on guilt, self-doubt, and psychological withdrawal following traumatic experiences. In contrast, \textit{No Stigma} posts are more oriented toward situational interpretation and clarification, including recollections of past sexual assault, impact of trauma on functioning and well-being, seeking clarification on sexual violence, and navigating ambiguous sexual encounters, indicating a focus on describing events and making sense of experiences rather than expressing \textit{Internalized Stigma}.

As shown in Table~\ref{tab:lloom_table_support}, comments on \textit{Stigma}-related posts emphasize validation and affirmation, including normalizing trauma responses, experience sharing, and empowering survivor identity, often alongside actionable guidance such as resource signposting and challenging harmful narratives. In contrast, comments associated with \textit{No Stigma} posts emphasize informational and guidance-oriented interactions, with themes such as affirming emotional support, empathetic validation, defining and identifying sexual assault-related concepts, and encouragement of professional mental health support. Compared to \textit{Stigma}-related responses, these comments place greater emphasis on clarification and practical advice rather than counteracting \textit{Internalized Stigma}.

To examine structural differences, we embedded LLooM-derived concepts and visualized them using dimensionality reduction. Figure~\ref{fig:concept_clusters}(a) shows partial separation between \textit{Stigma} and \textit{No Stigma} post concepts, with \textit{Stigma} clustering around internalized distress and \textit{No Stigma} around situational interpretation, indicating clear thematic differences. In contrast, Figure~\ref{fig:concept_clusters}(b) shows substantial overlap in comment-related concepts. While \textit{Stigma} support emphasizes validation and \textit{No Stigma} leans toward informational guidance, boundaries are less distinct, suggesting shared support strategies across contexts.

\begin{figure}[t]
    \centering
    \includegraphics[width=0.7\linewidth]{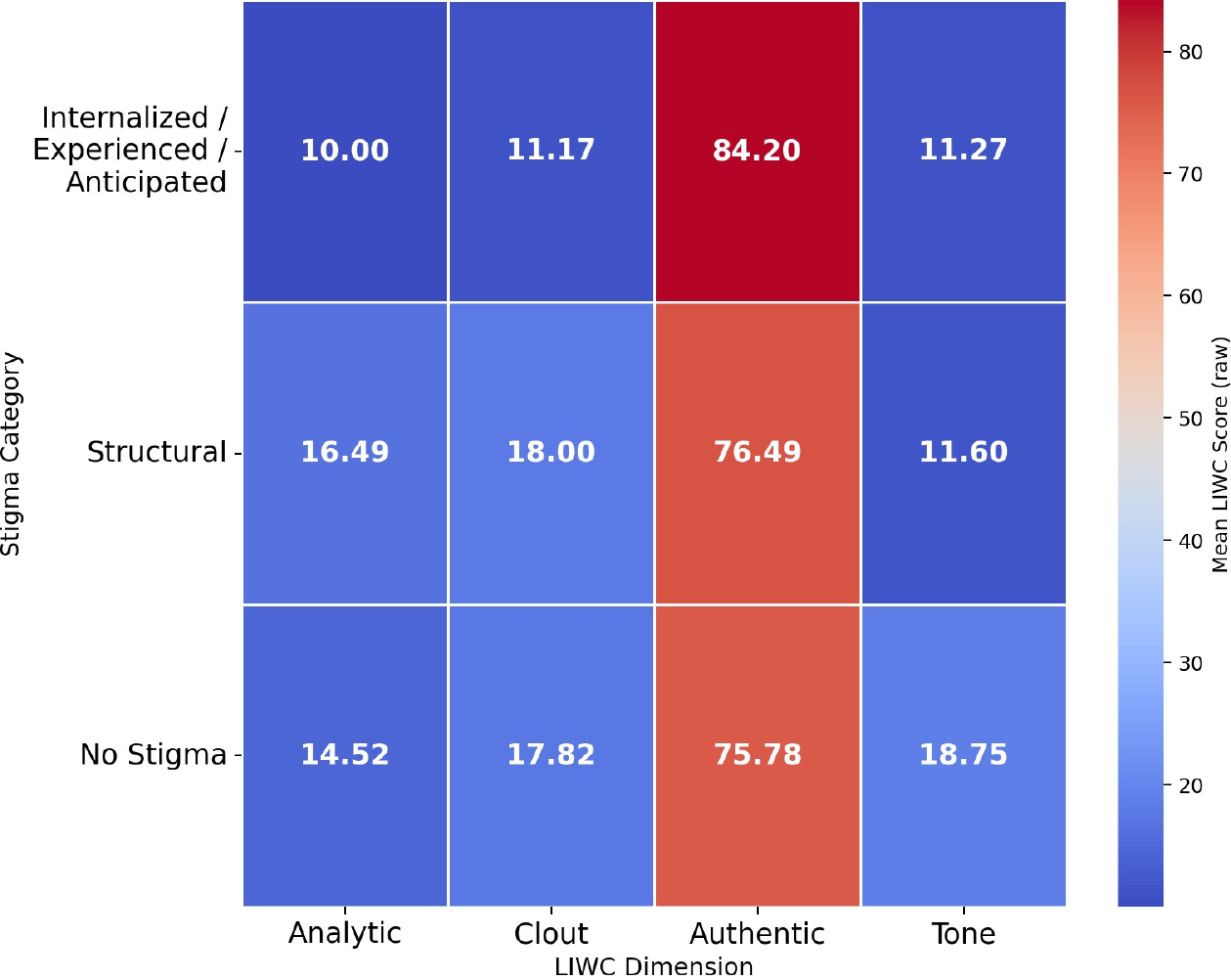}
    \caption{Post-level linguistic patterns across stigma.}
    \label{fig:liwc_d}
    \vspace{-0.25cm}
\end{figure}

\noindent\textbf{Linguistic Analysis Using LIWC. }  
We grouped stigma into two broad categories: \textit{Felt Stigma} (including \textit{Internalized, Experienced, Anticipated}), reflecting personal experiences, and \textit{Structural Stigma}, capturing systemic barriers. As shown in Figure~\ref{fig:liwc_d}, posts tagged as \textit{Internalized, Experienced}, and \textit{Anticipated} \textit{Stigma} exhibit higher \textit{Authenticity} (84.20) and lower \textit{Analytic} scores (10.00) than \textit{Structural} \textit{Stigma} (\textit{Authentic} = 76.49, \textit{Analytic} = 16.49), reflecting a more emotionally confessional style compared to the more analytical framing of structural barriers. \textit{No Stigma} posts fall between these extremes, suggesting a relatively neutral linguistic baseline.
At the comment level, \textit{Clout} scores are consistently high across all stigma categories and support types, indicating that responses generally adopt a confident, authoritative tone. \textit{Esteem Support} exhibits the highest \textit{Clout} across categories (76.26, 77.50, and 68.24 for \textit{Internalized/Experienced/Anticipated}, \textit{Structural}, and \textit{No Stigma}), suggesting that affirmation-based responses are particularly assertive. \textit{Group Interaction} shows the highest \textit{Authenticity}, reflecting the personal and experiential nature of peer engagement, while \textit{Information Support} has the highest \textit{Analytic} scores, consistent with its instructional and solution-oriented role.\footnote{see additional figures at \url{https://github.com/ShirleneRose/Stigma_SV/}}

\noindent\textbf{Emotion Analysis. }
To examine the emotional characteristics of \textit{Stigma} and \textit{No Stigma} content, we applied the NRC Emotion Lexicon, which captures eight basic emotions: anger, fear, sadness, disgust, anticipation, trust, joy, and surprise \citep{plutchik2001nature}. Across posts, narratives labeled as \textit{Internalized, Experienced, Anticipated Stigma} exhibit relatively higher proportions of negative emotions such as \textit{sadness}, \textit{fear}, and \textit{anger}, reflecting the distress and personal vulnerability associated with lived experiences of stigma. In contrast, \textit{Structural} posts show a more moderated emotional profile, with comparatively balanced levels of both negative and neutral emotions, indicating discussions that are less personal and more systemic in nature, while \textit{No Stigma} posts display higher levels of \textit{trust}, \textit{anticipation}, and \textit{joy}, suggesting more neutral or forward-looking conversations. A similar pattern is observed in support comments, where responses across all categories are dominated by \textit{trust} and \textit{anticipation}, but comments addressing \textit{Internalized/Experienced/Anticipated} content exhibit slightly higher negative affect, including \textit{sadness} and \textit{fear}, reflecting empathetic engagement with distressing experiences. Support for \textit{Structural} content remains comparatively neutral, while responses to \textit{No Stigma} posts maintain a more balanced and positive emotional distribution. Overall, these results demonstrate that emotionally intense signals are most prominent in personally \textit{Experienced Stigma}, whereas structural and \textit{No Stigma} content exhibit comparatively moderated emotional patterns.\footnote{See our GitHub repository for additional figures and results.}

\noindent\textbf{Support Dynamics in Survivor Narratives.} Our results show that support patterns are largely consistent across stigma categories, with differences mainly in emphasis rather than the categories employed. \textit{Information Support} is most prevalent, followed by \textit{Esteem} and \textit{Emotional Support}, reflecting a general focus on guidance and validation.
The differences are modest: \textit{Internalized, Experienced}, and \textit{Anticipated Stigma} receive slightly more \textit{Esteem Support} (26.1\%), \textit{Structural Stigma} shows higher \textit{Tangible Support} (7\%), and \textit{No Stigma} posts have the most \textit{Information Support} (38.7\%). Overall, communities rely on stable support strategies, adjusting emphasis based on the disclosure type.

\section{Related Work}
\label{sec:related_work}

Stigma has been widely conceptualized as a multidimensional social process involving labeling, stereotyping, and discrimination across social and institutional contexts \cite{link2001conceptualizing}, with subsequent work distinguishing mechanisms such as experienced, anticipated, and internalized stigma in health-related domains including HIV, mental illness, and substance use \cite{earnshaw2009conceptualizing, corrigan2012self, corrigan2005stigma, earnshaw2020stigma, smith2016substance}. While these frameworks establish stigma as a layered phenomenon affecting both individual perceptions and structural outcomes, they primarily focus on offline settings and do not examine how these mechanisms manifest in large-scale online discourse. Prior research on online platforms, particularly Reddit, shows that users frequently engage in self-disclosure and seek advice in anonymous environments \cite{andalibi2016understanding, de2014mental, garg2025analyzing, saxena2025trauma}, with responses often containing diverse forms of social support, including emotional, informational, and peer-based support expressed through linguistic patterns \cite{de2017language, sharma2018mental, sharma2020engagement}. Additional studies highlight structured interaction patterns in online health communication and community-driven platforms \cite{lopes2019classification, thukral2018analyzing}, but have not explicitly modeled how different types of stigma are embedded within narratives or how they relate to specific forms of support. 
Advances in computational methods, including natural language processing and large language models, have enabled large-scale analysis of stigmatizing language and thematic structures in online discourse \cite{bouzoubaa2024words, giorgi2024lived, straton2020stigma, lee2022mental, lam2024concept, Prompt-Engineering2025, LLM-ActiveLearning2023}, yet these approaches primarily focus on detection or representation learning rather than jointly capturing the multi-dimensional nature of stigma and its interaction with support. 
Our work addresses this gap through a hierarchical multi-label framework that links stigma expressions in survivor narratives to support strategies in community responses.

\section{Ethics and Limitations}
\label{sec:ethics}
This work examines sensitive discussions of sexual violence, so we take several steps to minimize potential harm. Although the data are publicly available on Reddit, posts may contain deeply personal experiences, and users may not expect them to be used for research. We therefore paraphrase all excerpts, remove potentially identifiable details, and only share post/comment IDs to avoid sharing the original content. The annotations capture expressions in the text and should not be interpreted as clinical or psychological assessments of individuals. In addition, the norms, moderation practices, and user populations of the selected subreddits may shape both survivor disclosures and community responses.
Our approach also has methodological limitations. LLM-based annotations may introduce systematic biases or miss nuanced and overlapping forms of stigma and support, and validation against manually annotated data reduces but does not eliminate these errors. The proposed taxonomy may not capture all expressions of stigma and support across diverse communities. Because the analysis is limited to English-language Reddit data, its generalizability may be constrained by demographic, platform-specific, and self-selection biases, and the findings may not reflect evolving patterns over time.
Finally, the dataset is imbalanced across stigma categories. Although this imbalance reflects the collected data, it limits comparisons for less common categories. We therefore interpret findings related to \textit{Structural Stigma} cautiously and call for larger, more balanced datasets in future work.

\section{Discussion and Conclusion}
\label{sec:conclusion}
This work provides an interaction-centered account of stigma and support in online sexual violence discussions. Prior research has established stigma as a multi-dimensional social process and has shown that Reddit can serve as a space for disclosure, anonymity, and support-seeking \cite{earnshaw2009conceptualizing, andalibi2016understanding, de2014mental, bouzoubaa2025phenotypes}. Our \corpus dataset links survivor narratives with the responses, enabling us to examine how disclosure framing and peer support are jointly shaped within socio-technical spaces.

A main finding of our work is the relative stability of community support norms. Across \textit{Stigma} and \textit{No Stigma} posts, commenters rely on a broadly consistent repertoire of \textit{Information}, \textit{Emotional Support}, and \textit{Esteem Support}, as well as \textit{Tangible Assistance} and \textit{Group Interaction}, with differences arising mainly in emphasis rather than in kind. This pattern suggests that community members draw on recognizable and durable response conventions that provide validation, guidance, and reassurance. Our findings extend prior work on online support exchange by showing that these response norms are resilient across different stigma profiles \cite{de2017language, sharma2020engagement, kraut2012building, chancellor2018norms}. This resilience also highlights how platform affordances, such as pseudonymous threading, can foster care even around deeply marginalized experiences \cite{dimond2013hollaback}.

The linguistic and emotional patterns further clarify why these support norms matter. Posts labeled as \textit{Internalized}, \textit{Experienced}, and \textit{Anticipated Stigma} are more confessional and affectively intense, whereas \textit{Structural Stigma} is framed in more analytical language. Community replies, however, tend to stabilize the interaction through validation, affirmation, and practical guidance. Theoretically, this suggests that online communities actively work to repair the ``identity'' that survivors anticipate or internalize \cite{goffmanStigmaNotesManagement1968, link2001conceptualizing}. Peer support does more than respond to distress; it helps make disclosures legible and credible. This aligns with prior work on reciprocity, linguistic accommodation, and community participation \cite{andalibi2018social, de2017language, sharma2018mental}. 

These results also suggest that moderation and support tools should preserve constructive responses such as validation, information, and belonging. More broadly, they show that platform outcomes depend on interaction norms as well as content \cite{gillespie2018custodians}. For automated and human-in-the-loop systems \cite{seering2019moderator}, the goal should not be to generate highly customized responses for every stigma subtype, but to leverage datasets like SCOPE to detect and amplify timely, safe, and norm-congruent forms of peer assistance.

In conclusion, SCOPE contributes a dataset and computational framework for studying stigma and support together rather than separately. The main contribution of this work is demonstrating that while stigma deeply shapes how survivors narrate their experiences, community responses remain structurally stable even as those narratives differ. This interactional perspective offers a foundation for future research on online sexual violence discourse and for designing safer, more responsive digital support environments across platforms with varying affordances, anonymity structures, and moderation regimes.

\begin{acks}
We thank Google for access to Gemini through the Gemini Academic Program Award. We also thank the Reddit communities whose public discussions made this work possible.
\end{acks}
\bibliographystyle{ACM-Reference-Format}

\bibliography{custom}


\end{document}